\documentclass[pagebackref,breaklinks,colorlinks,allcolors=cvprblue,10pt,twocolumn,letterpaper]{article}

\usepackage[accsupp]{axessibility}  

\usepackage[pagenumbers]{cvpr}

\usepackage{makecell}
\definecolor{cvprblue}{rgb}{0.21,0.49,0.74}
\def\paperID{3011} 
\def\confName{CVPR}
\def\confYear{2025}

\usepackage{tikz}
\newcommand{\blackcircle}[1]{%
    \begin{tikzpicture}[baseline=-0.9ex]
        \node[circle,fill=black,text=white,inner sep=0.2pt,minimum size=1pt] {#1};
    \end{tikzpicture}%
}

\definecolor{darkpurple}{RGB}{92, 0, 194}
\definecolor{darkblue}{RGB}{0, 66, 143}
\definecolor{darkgreen}{RGB}{66, 143, 24}
\definecolor{lightred}{RGB}{255, 204, 204}
\definecolor{lightblue}{RGB}{54, 125, 189}
\definecolor{green}{RGB}{0, 0, 0}

\newcommand{\SymbolicLang}{\textsc{$\mathcal{A}$-Language}\xspace}
\newcommand{\function}{\textbf{\textit{\textcolor{darkblue}{function}}}\xspace}
\newcommand{\parameter}{\textbf{\textcolor{darkgreen}{\textit{parameter}}}\xspace}
\newcommand{\topology}{\textbf{\textit{\textcolor{darkpurple}{topology}}}\xspace}

\usepackage{multirow}
\usepackage{fontawesome5}
\usepackage[T1]{fontenc}
\usepackage{pifont}
\usepackage{graphicx}  
\usepackage{float}
\usepackage{cuted}      
\usepackage{caption}    
\usepackage{tcolorbox}

\usepackage{hyperref}

\tcbuselibrary{listingsutf8}
\usepackage{listings}

\usepackage{colortbl}
\usepackage{tabularx}
\usepackage{array}
\usepackage{ulem}
\usepackage[capitalize,noabbrev]{cleveref}
\title{Symbolic Representation for Any-to-Any Generative Tasks}
\author{
    Jiaqi Chen$^{1,2,3}$\thanks{Equal Contribution.}~~\thanks{Corresponding author.}\hspace{1cm} 
    Xiaoye Zhu$^{4}$\footnotemark[1] \hspace{1cm}
    Yue Wang$^{5}$\footnotemark[1]\\ 
    Tianyang Liu$^6$ \hspace{0.6cm} 
    Xinhui Chen$^{7,8}$ \hspace{0.6cm} 
    Ying Chen$^9$ \hspace{0.6cm} 
    Chak Tou Leong$^{10}$ \hspace{0.6cm} 
    Yifei Ke$^8$ \\
    Joseph Liu$^{11}$ \hspace{0.6cm} 
    Yiwen Yuan$^{12}$ \hspace{0.6cm} 
    Julian McAuley$^{6}$
 \hspace{1cm}
    Li-jia Li$^{13}$ 
        \vspace{.5em}
    \\
    $^1$Stanford University  
    \quad
    $^2$Fellou AI 
    \quad
    $^3$Fudan University 
    \quad
    $^4$South China University of Technology 
    \quad \\
    $^5$Cornell University 
    \quad
    $^6$University of California San Diego
    \quad 
    $^7$Fenz.AI
    \quad
    $^8$Wuhan University
    \quad \\
    $^9$University of Illinois at Urbana-Champaign
    \quad 
    $^{10}$Hong Kong Polytechnic University
    \quad\\
    $^{11}$University of Southern California
    \quad 
    $^{12}$Carnegie Mellon University 
    \quad
    $^{13}$LiveX AI
      \vspace{.5em} 
  \\
  \textcolor{black}{\url{https://github.com/Jiaqi-Chen-00/Any-2-Any}}
}

\begin{document}

\maketitle

\begin{strip}
    \centering
    \vspace{-1cm}
    \includegraphics[width=\textwidth]{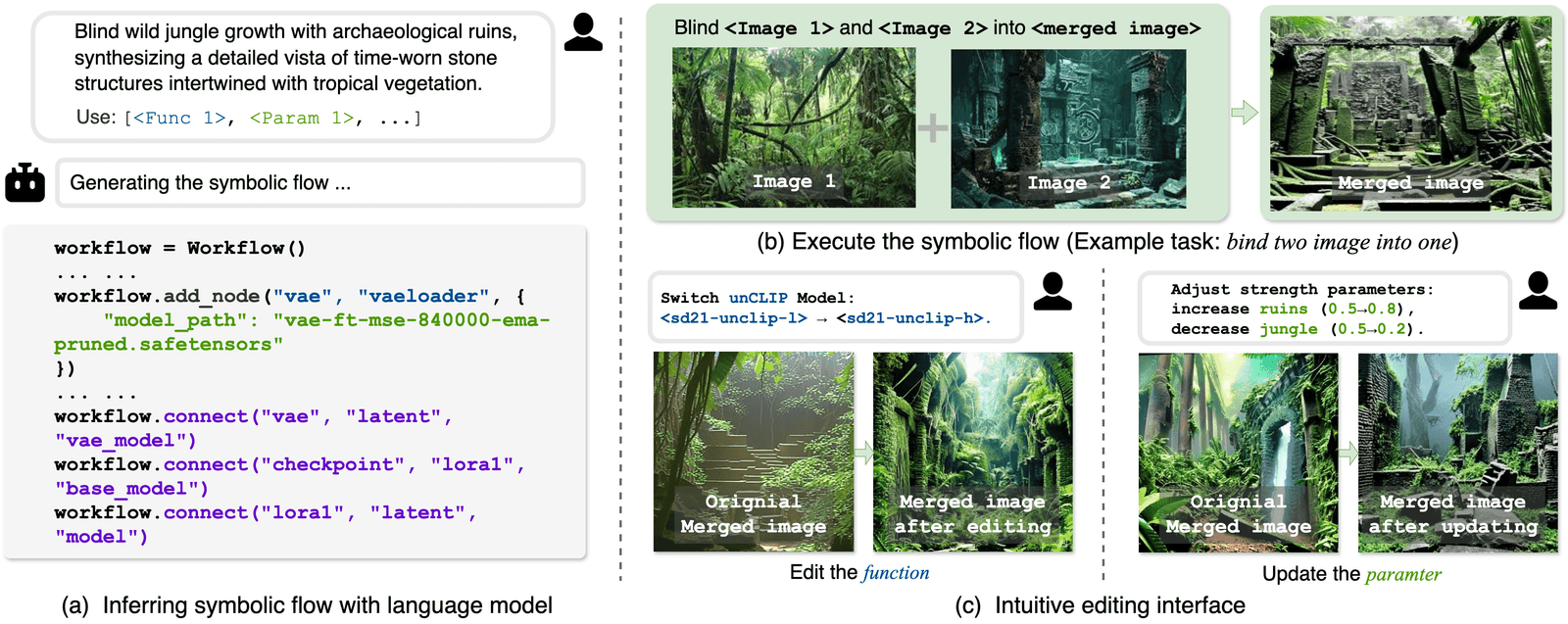}
    \captionof{figure}{
        \textbf{A symbolic representation for \textit{Any-to-Any} generative tasks.}
        (a) We develop a training-free inference engine that transforms natural language task descriptions into executable symbolic flow comprising \textcolor{darkblue}{\textit{functions}}, \textcolor{darkgreen}{\textit{parameters}}, and the \textcolor{darkpurple}{\textit{topology}}. 
        (b) The symbolic flow allows executing generative tasks as programs. Example task is mentioned in the first sentence of Sec.~\ref{sec:intro}. (c) Both \textcolor{darkblue}{\textit{functions}} and \textcolor{darkgreen}{\textit{parameters}} can be easily modified to customize the generation process and the output style.
    }
    \label{fig:intro}
\end{strip}


\begin{abstract}
We propose a symbolic generative task description language and a corresponding inference engine capable of representing arbitrary multimodal tasks as structured symbolic flows.  
Unlike conventional generative models that rely on large-scale training and implicit neural representations to learn cross-modal mappings—often at high computational cost and with limited flexibility—our framework introduces an explicit symbolic representation comprising three core primitives: \textcolor{darkblue}{functions}, \textcolor{darkgreen}{parameters}, and \textcolor{darkpurple}{topological logic}.  
Leveraging a pre-trained language model, our inference engine maps natural language instructions directly to symbolic workflows in a training-free manner.  
Our framework successfully performs over $12$ diverse multimodal generative tasks, demonstrating strong performance and flexibility without the need for task-specific tuning.  
Experiments show that our method not only matches or outperforms existing state-of-the-art unified models in content quality, but also offers greater efficiency, editability, and interruptibility.  
We believe that symbolic task representations provide a cost-effective and extensible foundation for advancing the capabilities of generative AI.
\end{abstract} 
\section{Introduction}
\label{sec:intro}
\begin{figure*}[h!]
    \centering
\includegraphics[width=\textwidth]{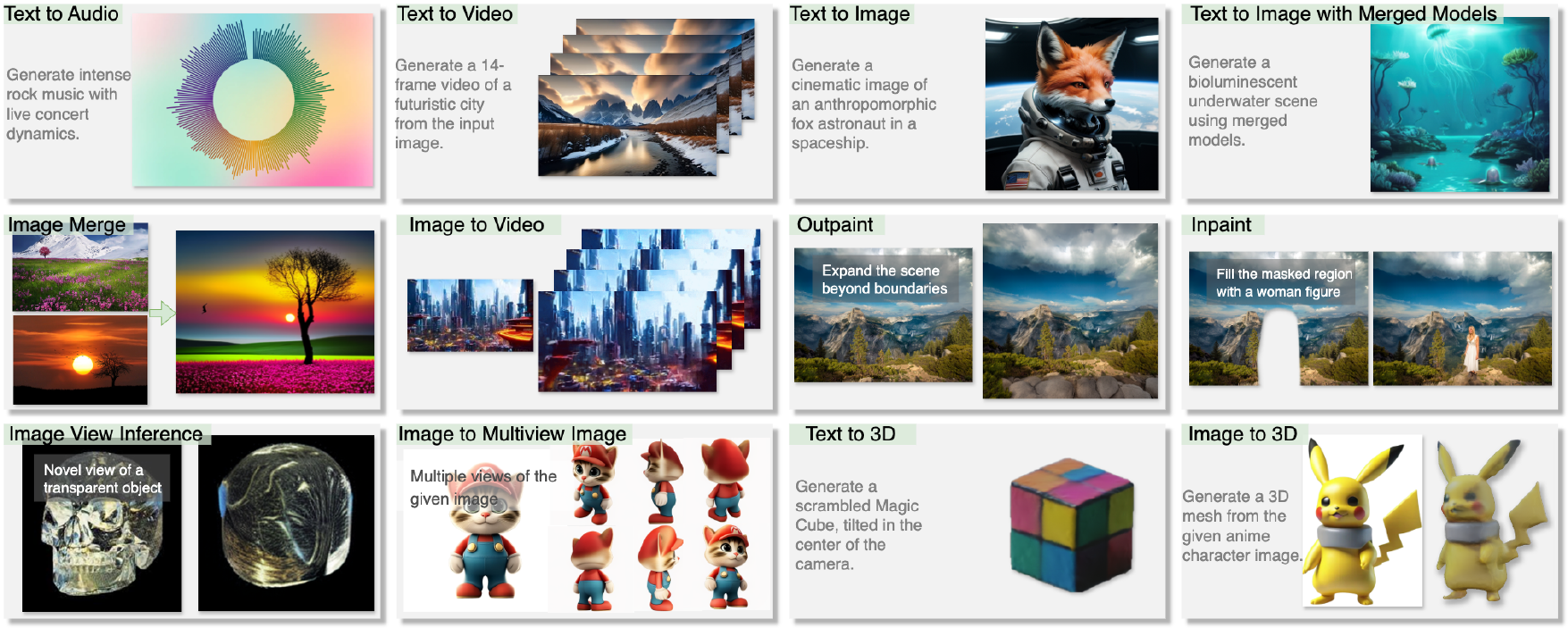}
    \caption{
\textbf{The \textit{Any-to-Any} generative model.}  Our model demonstrates the capability to handle \textbf{any-to-any generative tasks} across various modalities, including text, images, videos, audio, and 3D content. It supports flexible transformations such as converting image to video, generating 3D models from images, or synthesizing audio from textual prompts. Formally, any-to-any generative tasks refer to generating outputs in any desired modality from inputs in any other modality, all guided by natural language instructions~\cite{tang2024any}.}
    \label{fig:representation}
\end{figure*}
%

``\textit{Blending the wild growth of a jungle with the mystique of ancient ruins into a brand-new scene would be stunning},'' your artist friend mused. ``\textit{And if we could transform the photographic image into a video, overlayed with my audio recording of birds chirping and the soft murmur of flowing water—it would create a truly dreamlike sensory experience.}'' 
These increasingly complex, cross-modal creative needs point to a fundamental challenge: how can we design a \textit{unified model} capable of seamlessly handling generative tasks across any combination of input and output modalities (\textit{i.e.}, \textbf{\textit{any-to-any}} generative tasks, as shown in \Cref{fig:representation}), guided by natural language instructions~\cite{tang2024any,gupta2023visual,lu2022unifiediounifiedmodelvision,lu2023unifiedio2scalingautoregressive,xue2024genagent}?
Taking the example of blinding two photographic images (see Figure~\ref{fig:intro}), the workflow for executing this task comprises several essential processes~\cite{gupta2023visual,xue2024genagent,shen2023hugginggptsolvingaitasks}. 
First, the system imports two images and encodes them to extract their latent features. Then, taking these features as conditioning inputs, it combines them based on the user-specified blending strength and re-synthesizes the blended latent representation onto a blank latent canvas. Finally, the system decodes this latent representation into a viewable image.

Current approaches for any-to-any generative tasks typically fall into two paradigms: \textit{Implicit neural modeling} and \textit{agaentic approaches}. Implicit neural modeling approaches directly learn a neural representation from mass training data~\cite{zhan2024anygptunifiedmultimodalllm,tang2023anytoanygenerationcomposablediffusion, tang2023codi2incontextinterleavedinteractive, openai2024gpt4ocard, lu2023unifiedio2scalingautoregressive, lu2022unifiediounifiedmodelvision, lu2023unifiedio2scalingautoregressive}. While offering simplicity in representing multimodal information, their extensibility is constrained by the scope of the training data. 
They struggle to handle rare or unanticipated tasks—such as the image blending example in Figure~\ref{fig:intro}, if such cases are not accounted for during training. Moreover, their reliance on implicit neural representations makes them non-interruptible, leaving them ill-equipped to manage complex, multi-step workflows. 
%
Agentic approaches rely on sophisticated multi-agent coordination and tool orchestration~\cite{hao2024toolkengptaugmentingfrozenlanguage, schick2023toolformerlanguagemodelsteach, qin2023toolllm, lu2024chameleon, gupta2023visual, shen2023hugginggptsolvingaitasks}, which introduces system instability and operational overhead in their decision-making process. While powerful, these approaches lack a unified formal representation of tasks and fail to capture their inherent compositional nature. Our experiments reveal that complex agent designs do not necessarily outperform simpler ones, motivating us to explore an alternative direction: focusing on \textit{\textbf{unified task representations}} and \textit{\textbf{language model-friendly interfaces}} that enable direct task specification.

Examining the image-blending example reveals three fundamental components essential for executing generative tasks. At its core are distinct \textit{\textbf{\textcolor{darkblue}{functions}}} -computational operations such as image encoding, conditioning, and blending that transform inputs into desired outputs. Each function's behavior is shaped by \textit{\textbf{\textcolor{darkgreen}{parameters}}}, such as the blending strength and re-synthesis intensity, which fine-tune the operation to meet specific requirements. These functions do not operate in isolation; their \textit{\textbf{\textcolor{darkpurple}{topology}}}, or interconnected relationships, form a cohesive workflow that guides the progression from input to output. These three components, functions, parameters, and topology, together enable the effective execution of complex generative tasks.
Based on these insights, we propose \SymbolicLang, a formal representation that systematically captures these three essential components of generative tasks. 
In \SymbolicLang, \textcolor{darkblue}{\function} specifies the core computational operations, enabling the system to precisely identify and execute required transformations. 
\textcolor{darkgreen}{\parameter} provides fine-grained control over each operation's behavior, allowing users to adapt functions to specific task requirements. 
\textcolor{darkpurple}{\topology} formalizes the workflow structure, defining how functions interact and combine to accomplish complex generative goals. 
Through this three-component abstraction, \SymbolicLang enables flexible yet structured orchestration of generative tasks.

Alongside the symbolic generative task language, we introduce a \textit{\textbf{training-free inference engine}} that utilizes a pre-trained language model (LM) as its foundation to derive a symbolic representation from input instructions and a designated key function. 
Initially, the pre-trained LM identifies a comprehensive function set and parameter set from the natural language instruction, forming an initial functional and parametric structure. 
With this set of functions, we then predict the topology, outlining the dependencies among functions to form the complete symbolic representation. 
We also implement a refinement module, an iterative process activated upon any inference failure, enabling immediate corrections to resolve issues. 
Together, the \SymbolicLang, the inference engine, and the refinement module led to a high-quality system that provides flexible and precise workflow-building capabilities.

Experimentally, we constructed a dataset of $120$ real-world generative cases spanning $12$ task categories and validated the effectiveness of our approach through user studies and executability evaluations. The results demonstrate that our symbolic model is competitive with or outperforms state-of-the-art multimodal generative models in task generalization, output quality, and editing flexibility. Additionally, our experiments investigated the impact of syntax choices on the quality of symbolic flow generated by LMs.
Our contributions are three-fold:
\begin{itemize}
\item A unified symbolic representation, the \SymbolicLang, that systematically decomposes \textbf{any} generative task into three core components: \function for atomic operations, \parameter for behavioral control, and \topology for symbolic flow structure.
\item A \textbf{\textit{training-free inference engine}} that leverages pre-trained LMs to automatically convert natural language instructions into symbolic representations for executable workflows.
\item Empirical validation demonstrates its strong generalizability, modifiability, and user experience.
\end{itemize}
\section{Related work}
\label{rel_work}

\subsection{Unified multi-modal framework}

Recent years have witnessed remarkable advances in large language models (LLMs), which have demonstrated exceptional capabilities across various natural language tasks, from basic comprehension to complex reasoning~\cite{brown2020language,chowdhery2023palm, li2023starcoder, touvron2023llama,touvron2023llama2,chatgpt,openai2024gpt4, deepseekai2024deepseek, lozhkov2024starcoder,openai2024gpt4ocard,Claude2024Anthropic, jiang2024mixtralexperts}. Building on this success, multimodal large language models (MLLMs) have extended these capabilities to integrate multiple forms of input and output, covering data modalities such as images, audio, video, and 3D structures~\cite{li2020oscar,akbari2021vatt,fang2021clip2video,yan2021video,li2021align,radford2021learning,li2022blip,zellers2022merlot,zeng2022multi,yang2022vision,wang2022ofa,wang2022image, runway2024gen3, radford2022robustspeechrecognitionlargescale, borsos2023audiolmlanguagemodelingapproach, zhang2023speechgptempoweringlargelanguage, hong2022cogvideolargescalepretrainingtexttovideo, kreuk2023audiogentextuallyguidedaudio, rombach2022highresolutionimagesynthesislatent, BetkerImprovingIG, poole2022dreamfusiontextto3dusing2d}. The field has progressed from isolated single-modality models to sophisticated any-to-any frameworks~\cite{tang2023anytoanygenerationcomposablediffusion, tang2023codi2incontextinterleavedinteractive, zhan2024anygptunifiedmultimodalllm, openai2024gpt4ocard, lu2022unifiediounifiedmodelvision, lu2023unifiedio2scalingautoregressive, mizrahi20234mmassivelymultimodalmasked} that can handle diverse input-output combinations within a single model architecture.
However, these unified multimodal frameworks face significant challenges in practice. The scarcity of high-quality, diverse multimodal datasets remains a fundamental bottleneck, particularly for complex cross-modal tasks. Moreover, different modalities often require distinct processing approaches and representations, making it challenging to achieve optimal performance across all possible modality combinations in a single model. The need to align disparate modalities into a coherent unified representation while preserving their unique characteristics continues to be a core challenge in advancing these frameworks.


\subsection{Workflow synthesis}
Workflow synthesis~\cite{hu2017learning, johnson2017inferring, andreas2017neuralmodulenetworks} seeks to generate executable sequences of operations for complex tasks by coordinating AI models and resources, particularly in generative AI, where tasks often require sophisticated combinations of inference, parameters, and logic. Traditional methods using neural modules or predefined operations struggle with the open-ended nature of modern AI tasks. Recent advances like HuggingGPT~\cite{shen2023hugginggptsolvingaitasks} leverage large language models for task planning and model coordination, VISPROG~\cite{gupta2023visual} employs neuro-symbolic approaches for programmatic task decomposition, and GenAgent~\cite{xue2024genagent} uses multi-agent collaboration to build workflows step by step. Despite their differences, these approaches highlight the need for flexible, interpretable representations. Our work advances this field by proposing a unified symbolic framework for describing and executing generative tasks, balancing expressiveness and practicality.

\section{\texorpdfstring{$\mathcal{A}$}{A}-Language}
\label{sec:method}
We introduce \SymbolicLang, a symbolic representation that bridges the gap between natural language task descriptions and executable workflows for any-to-any generative tasks. Unlike previous unified multimodal approaches dependent on implicit neural representations and intensive training, our \SymbolicLang provides an \textbf{\textit{explicit symbolic representation}} (Sec.~\ref{sec:formulation} and~\ref{sec:syntax}), allowing a \textbf{\textit{training-free}} execution (Sec.~\ref{sec:inferring}).

\subsection{Formulation}
\label{sec:formulation}
Fundamentally, \SymbolicLang formalizes any generative task $t$ as a triple:
$$
    \Omega(t) := (\mathcal{F}, \Phi, \mathcal{T}).
$$
This unified formulation decomposes any generative task into its essential constituents: the computational \textcolor{darkblue}{\textbf{\textit{functions}}} $\mathcal{F}$, their corresponding \textcolor{darkgreen}{\textbf{\textit{parameters}}} $\Phi$, and the \textcolor{darkpurple}{\textbf{\textit{topological structure}}} $\mathcal{T}$ that elucidates their interrelations and data flow dynamics.

\paragraph{Function}
The function set is defined as $\mathcal{F} = \{f_1, f_2, ..., f_n\}$, where $n \in \mathbb{N}$, which represents atomic computational units. Each function takes both input data and parameters to produce outputs, formally defined as:
$$
f_i: \mathcal{I}_i \times \phi_i \rightarrow \mathcal{O}_i,
$$
where $\mathcal{I}_i$ defines its input space, $\phi_i$ represents its parameter configuration, and $\mathcal{O}_i$ specifies its output space. The input and output spaces $\mathcal{I}_i$ and $\mathcal{O}_i$ represent either simple scalar values or composite data structures of arbitrary modalities, allowing functions to process multiple inputs and generate multiple outputs. For example, an image blending function might accept two image inputs and produce both a blended result and an attention mask. When functions are connected, their inputs and outputs can be partially mapped, providing flexibility in constructing complex paths.

\paragraph{Parameter}
The parameter space $\Phi = \{\phi_{f_1}, \phi_{f_2}, ..., \phi_{f_n}\}$ encompasses configurations that modify function behaviors, where each $\phi_{f_i}$ represents the parameter space for function $f_i$. Parameters must be fully specified before function execution to ensure deterministic behavior. The parameter space is independent of the input space, enabling functions to exhibit different behaviors while processing identical inputs. 

\paragraph{Topology}
The topology set \( \mathcal{T} = \{d_1, d_2, ..., d_m\} \) defines the precise data flows between functions, where each \( d_k \) at the finest granularity specifies a single directed connection from a specific output of one function to a specific input of another function. Specifically, \( d_k \) is defined as a tuple representing an individual data flow from the output of a source function to the input of a target function. Formally:
$$
d_k = (f_j, y_j) \rightarrow  (f_i, x_i) \mid y_j \in \mathcal{O}_j,  x_i \in \mathcal{I}_i
$$
where \( f_j \) and \( f_i \) denote the source and target functions, respectively. \( y_j \) refers to a specific output produced by function \( f_j \), while \( x_i \) corresponds to a specific input required by function \( f_i \). Thus, each \( d_k \) encapsulates the transfer of data from a designated output of one function to a designated input of another, allowing for precise tracking of data flow through the system. 

\paragraph{Symbolic flow}
The symbolic flow emerges from the interaction of \textcolor{darkblue}{\textit{\textbf{functions}}}, \textcolor{darkgreen}{\textit{\textbf{parameters}}}, and \textcolor{darkpurple}{\textit{\textbf{topological logic}}}, formalizing the complete generative process:
$$
\mathcal{S} = \{(f_i, \phi_{f_i}, D_i) \mid f_i \in \mathcal{F}\},
$$
where \( D_i \) is the set of all data flows \( d_k \) in \( \mathcal{T} \) that target function \( f_i \):
$$
D_i = \{ (f_j, y_j) \rightarrow (f_i, x_i) \mid f_j \in \mathcal{F}, y_j \in \mathcal{O}_j, x_i \in \mathcal{I}_i \}.
$$
Each element in the symbolic flow specifies a function, its parameter configuration, and its incoming directed connections. Specifically, for each function \( f_i \), \( D_i \) contains tuples that map specific outputs of predecessor functions to specific inputs of \( f_i \). This fine-grained formulation captures how computation progresses through the system, with functions receiving their required inputs from designated outputs of antecedent functions and parameter configurations from the parameter space. Through this unified and detailed representation, \SymbolicLang can express diverse and complex generative tasks.
\begin{table*}[t]
\centering
\footnotesize
\setlength{\tabcolsep}{12pt}
\begin{tabular}{ll}
\toprule
\textbf{Notation} & \textbf{Implementation and definition} \\
\midrule
\underline{\textit{\textbf{System Components}}} \\[0.5em]
$\mathcal{X}$ & \texttt{List[Any]} // Input data of any modality \\
$s$ & \texttt{str} // Task description \\
$\mathcal{C}$ & \texttt{Dict} // System constraints \\
$\Omega(t)$ & \texttt{Workflow} // Complete workflow representation \\
\midrule
\underline{\textit{\textbf{Workflow Structure}}} \\[0.5em]
$f_i \in \mathcal{F}$ & \texttt{Node} // Computational function \\
$f_i: \mathcal{I}_i \times \phi_i \rightarrow \mathcal{O}_i$ & \texttt{Node.forward} // Function mapping with parameters \\
$\phi_{f_i} \in \Phi$ & \texttt{Dict[str, Any]} // Function parameters \\
$d_k \in \mathcal{T}$ & \texttt{(Node, Any)-> (Node, Any)} // Source output to target input mapping ($(f_j, y_j) \rightarrow (f_i, x_i)$) \\
\midrule
\underline{\textit{\textbf{Workflow Operations}}} & \textit{(Declarative syntax, simplified version)}\\[0.5em]
Initialize & \texttt{Workflow()} // Create empty workflow $\Omega(t) = (\mathcal{F}, \Phi, \mathcal{T})$ \\
Add Node & \texttt{add\_node(name, type, params)} // Add function $f_i$ with parameters $\phi_{f_i}$ \\
Connect & \texttt{connect(src\_node, src\_output, dst\_node, dst\_input)} // Create topology $d_k: (f_j, y_j) \rightarrow (f_i, x_i)$ \\
\bottomrule
\end{tabular}
\caption{\textbf{System components and operations summary.} A comprehensive overview of \SymbolicLang's system components and their implementations. The upper two sections define the mathematical notations and their corresponding implementations, where the system processes input data $\mathcal{X}$ according to task description $s$ under constraints $\mathcal{C}$. Functions $f_i$ transform inputs $\mathcal{I}_i$ with parameters $\phi_i$ to outputs $\mathcal{O}_i$, and are connected through directed mappings $d_k$. The lower section demonstrates the Declarative Syntax as one example of workflow construction, showing how basic operations map to the mathematical formulation $\Omega(t) = (\mathcal{F}, \Phi, \mathcal{T})$.}
\vspace{-1.5em}
\label{tab:system_summary}
\end{table*}

\subsection{Syntax styles}\label{Syntax_styles}
\label{sec:syntax}
\begin{figure}[t]
    \centering
\includegraphics[width=\linewidth]{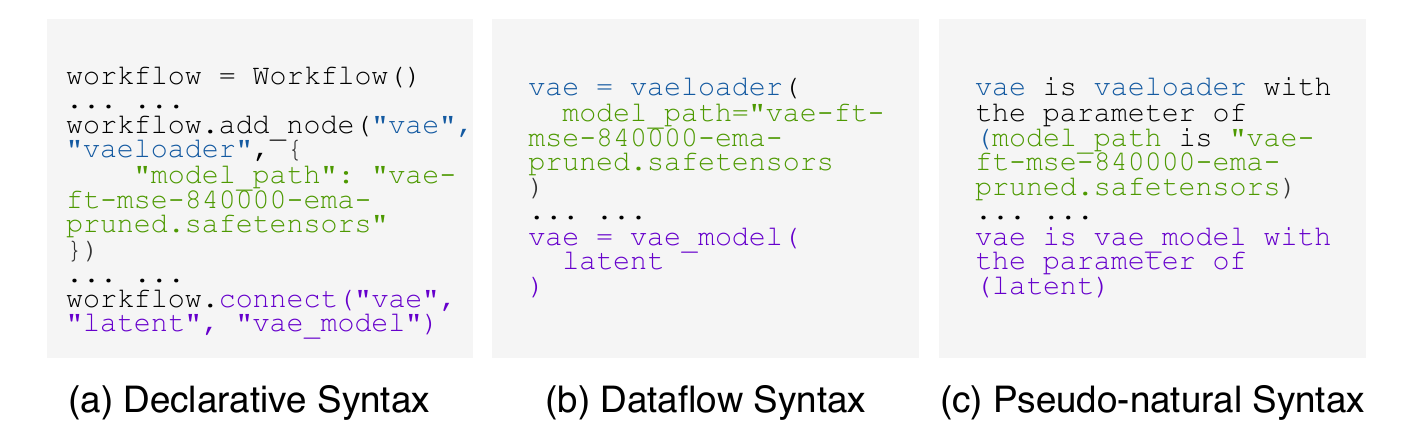}
    \caption{
\textbf{Syntax comparison.}
We implement our symbolic representation using three different styles of domain-specific languages (DSLs). (a) The declarative syntax registers all components into the workflow. (b) The dataflow syntax emphasizes the direction of data flow. (c) The pseudo-natural syntax mimics human language expression.
}
    \vspace{-1.5em}
    \label{fig:syntax_comparison}
\end{figure}

The symbolic representation $\Omega(t)$ can be expressed through multiple syntactic styles, as shown in Figure~\ref{fig:syntax_comparison}, each offering different trade-offs in expressiveness and clarity. 
To identify the most effective representation for large language model inference, we explore three distinct syntactic formulations: \textit{\textbf{declarative}}, \textit{\textbf{dataflow}}, and \textit{\textbf{pseudo-natural}} syntax, as illustrated through concise examples in Figure~\ref{fig:syntax_comparison}.

\paragraph{Declarative Syntax}
Declarative syntax~\cite{ullman2001neurocognitive} focuses on explicitly specifying computational components and their relationships. Functions are separately declared with parameters, while connections are specified through explicit statements. This style is effective for complex workflows with reusable components, as it clearly separates component definitions ($\mathcal{F}$) from relationships ($\mathcal{T}$).

\paragraph{Dataflow syntax}
Dataflow syntax~\cite{xue2024genagent} emphasizes the flow of data through function compositions, where outputs directly feed into subsequent functions. It captures topological relationships ($\mathcal{T}$) through the order of function calls while maintaining explicit parameter specifications ($\Phi$). This style is particularly suited for linear, sequential workflows.

\paragraph{Pseudo-natural syntax}
Pseudo-natural syntax~\cite{ernst2017natural} aims to bridge formal representations with more intuitive, language-like structures, making task specifications more accessible while maintaining mathematical rigor. This style explores a balance between precision and readability.

Each style retains the full expressiveness of $\Omega(t)$, but offers different advantages in terms of clarity and usability. The subsequent empirical analysis will evaluate which syntax best supports natural language inference while preserving necessary formal properties.

\section{Inferring via pre-trained language model}
\label{sec:inferring}
\begin{figure*}[h]
    \centering
    \includegraphics[width=\textwidth]{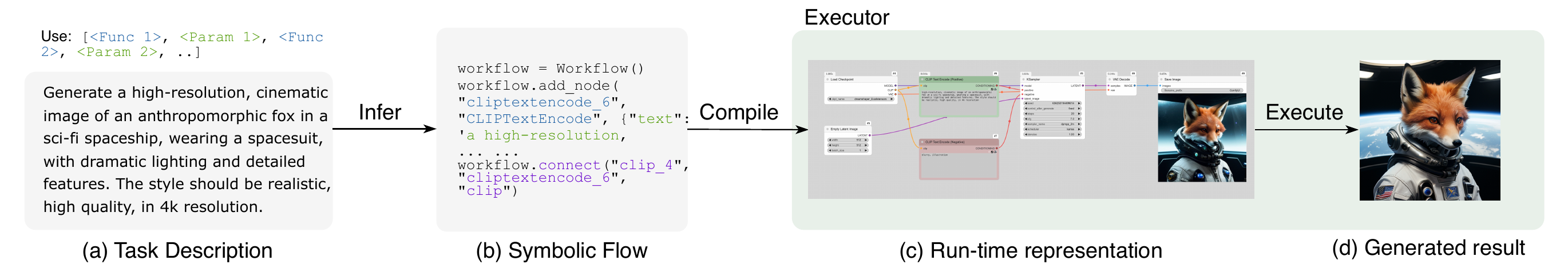}
    \caption{\textbf{Demonstration of the inference and execution.} The inference framework translates a natural language task description into an executable symbolic representation. This symbolic representation is then compiled and executed through a workflow executor to perform the desired transformation. See appendix for details.}
    \vspace{-1.5em}
\label{fig:symbolic_workflow_generation}
\end{figure*}

The diversity and complexity of generative tasks necessitate a flexible and robust approach to transforming high-level task specifications into executable symbolic flows.  As illustrated in Figure~\ref{fig:inference}, we propose utilizing LMs as inference engines to generate task-specific symbolic representations, with Figure~\ref{fig:symbolic_workflow_generation} demonstrating the complete pipeline from natural language description to executable workflow. This enables any-to-any transformations across different modalities and task types.

Given a set of inputs $\mathcal{X}$ of arbitrary modalities, a task description $s$, and a set of constraints $\mathcal{C}$, our inference framework generates a complete symbolic representation $\Omega(t)$. As illustrated in Figure~\ref{fig:inference}, our framework leverages a pre-trained language model to infer both the computational components and their topology from natural language descriptions. This process can be formalized as:
$$
    \mathcal{M}: (\mathcal{X}, s, \mathcal{C}) \rightarrow \Omega(t),
$$
where $\mathcal{X}$ represents any combination of inputs such as images, text, audio, or other modalities, $s$ describes the desired transformation, and $\mathcal{C}$ represents a set of constraints, which typically specifying information such as available functions, specific parameter choices, valid parameter ranges, and model compatibility. These constraints are essential for ensuring that the generated symbolic flow is not only theoretically sound but also practically executable within the given computational environment. Specifically, we divide the inference into three main steps:
\begin{figure*}[h!]
    \centering
\includegraphics[width=\textwidth]{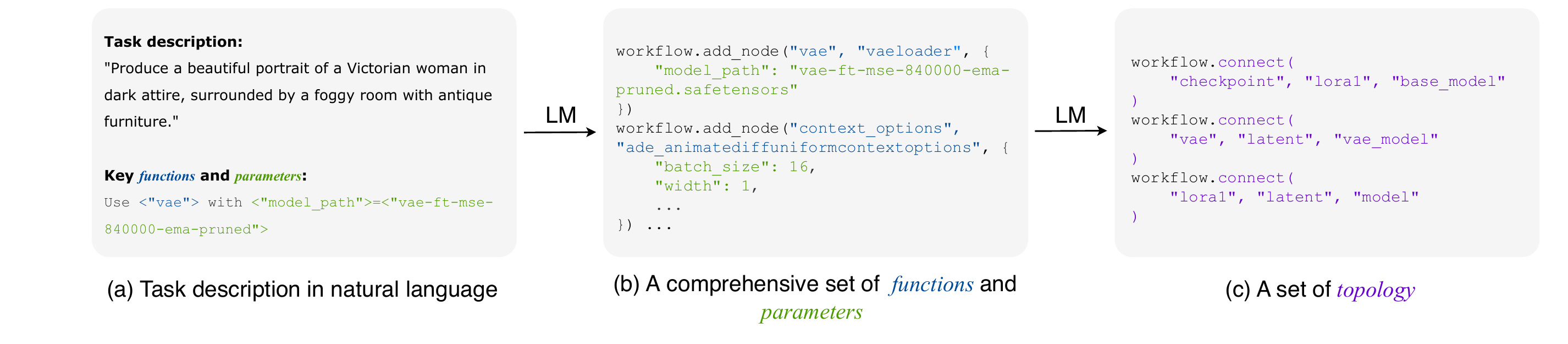}
    \caption{
\textbf{Inferring symbolic flow with pre-trained language model (LM).}
Beginning with (a) a natural language task description and key functions and parameters, we leverage LM to infer (b) a comprehensive set of functions and parameters. We then integrate (a) and (b) to deduce the (c) topology. If compilation or execution fails, all information is aggregated for further refinement (Sec.~\ref{sec:topology_construction}).
    }
    \label{fig:inference}
\end{figure*}

\paragraph{Component inference}
The first stage of our framework focuses on determining the necessary computational components. Given the input specifications and constraints, the LM identifies the required functions and their parameters:
$$
    \psi_1: (\mathcal{X}, s, \mathcal{C}) \rightarrow (\mathcal{F}, \Phi).
$$
This process accounts for both the explicit requirements of the task and any implicit dependencies, ensuring that selected functions are available within $\mathcal{C}$.

\paragraph{Topology construction}
\label{sec:topology_construction}
The second stage focuses on establishing relationships between the identified components to form a coherent computational flow:
$$
    \psi_2: (\mathcal{X}, s, \mathcal{C}, \mathcal{F}, \Phi) \rightarrow \mathcal{T}.
$$
In this phase, the LM evaluates how the outputs of one function can serve as inputs to another, ensuring that these connections are executable and comply with the constraints defined in $\mathcal{C}$. This construction guarantees that data flows seamlessly through the system in a manner consistent with our unified formulation.

\paragraph{Iterative refinement}
The generated symbolic flow undergoes an iterative refinement process to ensure correctness and executability. We define this refinement as:
$$
    \Omega_{i+1}(t) = R(\Omega_i(t), \epsilon_i),
$$
where $R$ represents the refinement operator and $\epsilon_i$ captures any detected issues in iteration $i$. To prevent endless loops, a maximum number of iterations can be set. During each iteration, the LM analyzes error signals and adjusts the symbolic flow accordingly, either by modifying function parameters, adding missing components, or restructuring topological connections. This iterative process continues until a valid symbolic flow is achieved that satisfies all constraints in $\mathcal{C}$ or the maximum iteration count is reached.

The combination of LM-based inference and iterative refinement enables our framework to handle diverse transformation tasks while maintaining robustness and generality. By leveraging the LM's reasoning capabilities and incorporating explicit constraints, we bridge the gap between high-level task descriptions and executable symbolic flows, providing a flexible foundation for any-to-any transformations.

\section{Experiments}
\label{sec:exp}

\subsection{Setup}
\paragraph{Evaluation Benchmarks}
We comprehensively evaluated our symbolic approach using 2 benchmarks: \blackcircle{1} A diverse task suite with $120$ generative tasks from real-world applications, categorized into $12$ general groups with $10$ instances each (see Appendix for the complete task list). \blackcircle{2} ComfyBench~\cite{xue2024genagent}, containing $200$ multi-step generative task workflows that integrate multiple components. 

\paragraph{Metrics}
For execution evaluation, we measured the single-run \textit{\textbf{pass rate (Pass@$\textbf{1}$)}} of compilation and execution on our task suite, and the \textit{\textbf{resolve rate}} on ComfyBench representing successful task completion. For outcome quality and instruction-following, we conducted a systematic user study with five annotators who ranked outputs from all frameworks using metrics including: \textit{\textbf{task-outcome alignment}} (correspondence between outputs and task specifications), \textit{\textbf{outcome quality}} (aesthetic appeal, structural coherence, and technical quality), \textit{\textbf{average rank}} (mean performance ranking across tasks), and \textit{\textbf{win rate}} (percentage of comparisons where our method ranked higher).

\paragraph{Baselines}
For our diverse task suite evaluation, we primarily compared with GenAgent~\cite{xue2024genagent} as our agentic framework baseline, augmented with key functions and up to $3$ refinement iterations for fairness. We also compared against unified multimodal models including Show-o~\cite{xie2024showo} (guidance scale $1.75$, $16$ time steps), SEED-x~\cite{ge2024seed} (maximum $1024$ tokens, $3$ history rounds), LVM~\cite{liu2023world}, and Unified-IO~\cite{lu2023unifiedio2scalingautoregressive}. For video generation, we included the commercial Gen-3~\cite{runway2024gen3} model ($720$p resolution, $5$-second length). For ComfyBench evaluation, we compared with training-free approaches (HuggingGPT~\cite{shen2023hugginggptsolvingaitasks} and ComfyAgent~\cite{xue2024genagent}) and the MLVM-based LWM~\cite{liu2023world} approach. 

\paragraph{Implementation details}
Following Gupta~\textit{et al.}~\cite{gupta2023visual}, we implemented in-context learning with syntax and logical guidance. We performed Retrieval-Augmented Generation based on task descriptions, retrieving three most relevant programs from a curated database containing $16$ distinct reference programs. All experiments ran on a single L4 GPU ($24$GB) with 1TB storage on a Debian 11 server. ComfyUI served as the back-end for code execution. We used GPT-4o (\texttt{gpt-4o-2024-08-06}) as the inference engine and \texttt{text-embedding-3-large} as the embedding model.

\subsection{Main results}


\begin{table}[htbp]
    \centering
    \scriptsize
    \setlength{\tabcolsep}{2.2pt}
\renewcommand\arraystretch{1.2}
    \caption{\textbf{Comparison of the average rankings} between outcome quality and task-outcome alignment rankings (↓) on our task suite.  We primarily compared \textit{\textbf{neural representing, training-dependent modeling}}~\cite{xie2024showo,ge2024seed,liu2023world,lu2023unifiedio2scalingautoregressive} and our \textit{\textbf{symbolic representing, training-free modeling}}. Each method was ranked on a scale starting from $1$, with $1$ denoting the best-performing approach. 
    ``U-IO 2'' denotes ``Unified-IO 2'', ``I-2-3D'' denotes ``Image to 3D Mesh'', ``T2M'' denotes ``Text to Mesh''.
    }
    \label{tab:quality_rank_comparison}
    \begin{tabular}{l|cccccccccccccc}
    \hline
    \toprule
    \textbf{Method} & \textbf{Inpaint} & \textbf{Outpaint} & \textbf{Img merge} & \textbf{NVS} & \textbf{Merge model} & \textbf{I-2-3D} \\
    \midrule
    Show-o~\cite{xie2024showo} & 1.6 & \textbf{1.4}  & \textcolor{lightred}{\ding{55}}  & \textcolor{lightred}{\ding{55}} & \textcolor{lightred}{\ding{55}} & \textcolor{lightred}{\ding{55}} \\
    SEED-X~\cite{ge2024seed} & \textcolor{lightred}{\ding{55}}    & \textcolor{lightred}{\ding{55}} & \textbf{1.2}    & \textcolor{lightred}{\ding{55}} & \textcolor{lightred}{\ding{55}}  & \textcolor{lightred}{\ding{55}}  \\
    LWM~\cite{liu2023world} & \textcolor{lightred}{\ding{55}}     & \textcolor{lightred}{\ding{55}}    & \textcolor{lightred}{\ding{55}}     & \textcolor{lightred}{\ding{55}}  & \textcolor{lightred}{\ding{55}}  & \textcolor{lightred}{\ding{55}}  \\
    U-IO 2~\cite{lu2023unifiedio2scalingautoregressive} & -    & \textcolor{lightred}{\ding{55}} & - & \textcolor{lightred}{\ding{55}}  & \textcolor{lightred}{\ding{55}}  & \textcolor{lightred}{\ding{55}}  \\
    \rowcolor[gray]{.9} 
    Ours & \textbf{1.4} & 1.6 & 1.8 & \textbf{1.0} & \textbf{1.0} & \textbf{1.0} \\
    \hline
    \hline
    \textbf{Method} & \textbf{T2I} & \textbf{T2A} & \textbf{Multi-view img} & \textbf{I2V}  & \textbf{T2M} & \textbf{T2V} \\
    \midrule
    Show-o~\cite{xie2024showo}& \textcolor{green}{2.8} & \textcolor{lightred}{\ding{55}}  & \textcolor{lightred}{\ding{55}}     & \textcolor{lightred}{\ding{55}}     & \textcolor{lightred}{\ding{55}}  & \textcolor{lightred}{\ding{55}}  \\
    SEED-X~\cite{ge2024seed} & \textcolor{green}{2.0} & \textcolor{lightred}{\ding{55}} & \textcolor{lightred}{\ding{55}}    & \textcolor{lightred}{\ding{55}}       & \textcolor{lightred}{\ding{55}} & \textcolor{lightred}{\ding{55}} \\
    LWM~\cite{liu2023world} & \textcolor{green}{4.2} & \textcolor{lightred}{\ding{55}} & \textcolor{lightred}{\ding{55}}    & \textcolor{lightred}{\ding{55}}     & \textcolor{lightred}{\ding{55}} & \textcolor{lightred}{\ding{55}} \\
    U-IO 2~\cite{lu2023unifiedio2scalingautoregressive} & \textcolor{green}{4.5} & \textcolor{green}{2.0}  & - & - & \textcolor{lightred}{\ding{55}} & \textcolor{lightred}{\ding{55}} \\
    \rowcolor[gray]{.9} 
    Ours & \textbf{\textcolor{green}{1.5}}  & \textbf{\textcolor{green}{1.0}} & \textbf{\textcolor{green}{1.0}} & \textbf{\textcolor{green}{1.0}}& \textbf{\textcolor{green}{1.0}} & \textbf{\textcolor{green}{1.0}} \\
    \hline

    \hline
    \end{tabular}%
   \vspace{-0.5cm}
    \end{table}

\begin{figure}[t]
    \centering
\includegraphics[width=\linewidth]{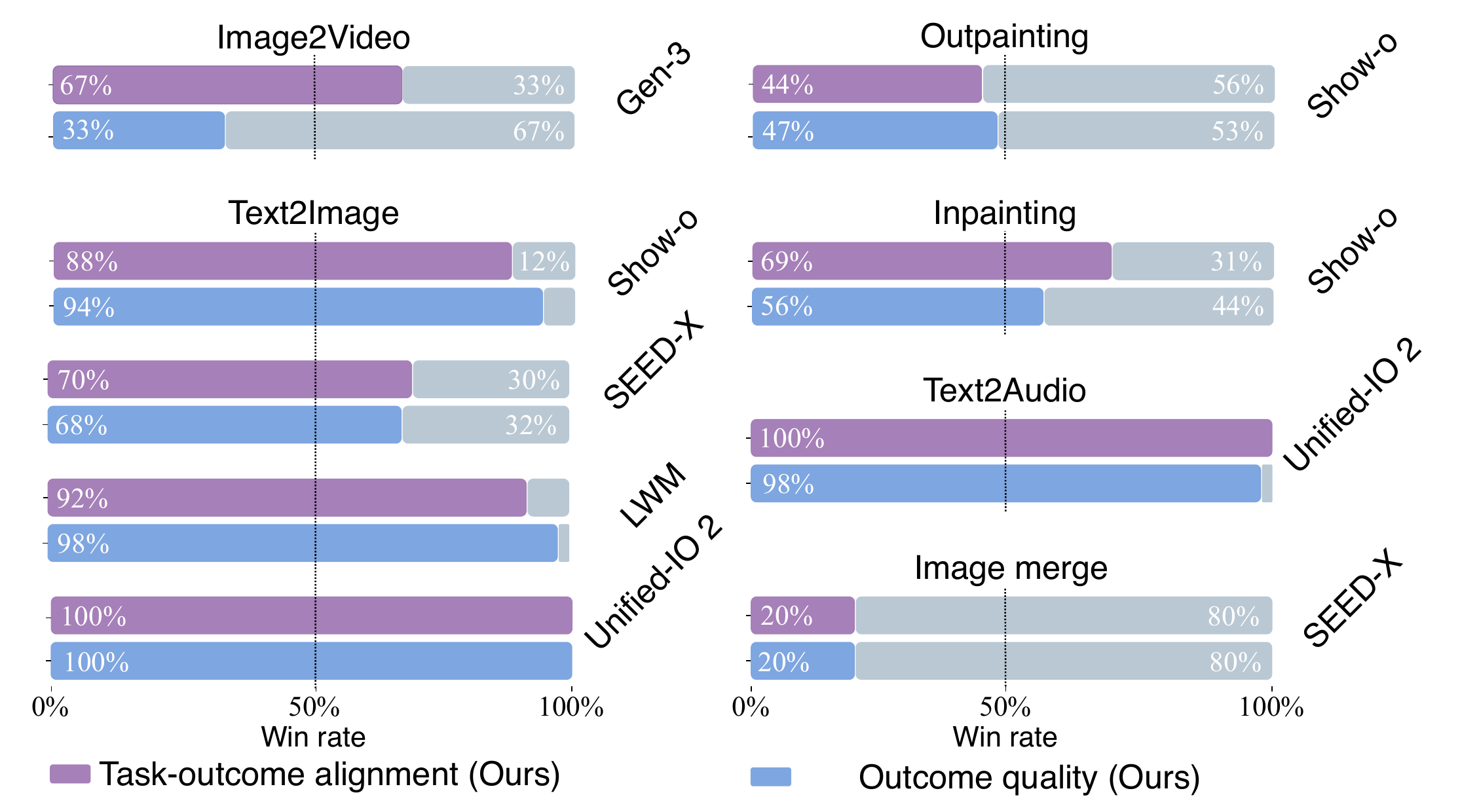}
    \caption{
\textbf{Comparison of our win rates} with the state-of-the-art unified multimodal models on our task suite.}
    \label{fig:baselines}
\end{figure}

\paragraph{Comparative performance in user study}
Our symbolic model consistently outperforms state-of-the-art unified models in both text-outcome alignment and result quality. As illustrated in Figure~\ref{fig:baselines}, our approach achieved a $94\%$ win rate against Show-o and $98\%$ against LVM in Text to Image tasks. In Image2Video generation, our model surpassed the commercial Gen-3 with a $67\%$ win rate in text-outcome alignment. For Text to Audio, our model attained a $100\%$ win rate in alignment and $98\%$ in quality against Unified-IO, underscoring its superior performance across diverse applications.

\begin{table}[t]
\centering
\raggedright
\small
\renewcommand\arraystretch{1.2}
\setlength{\tabcolsep}{5pt}
    \caption{\textbf{Performance on ComfyBench~\cite{xue2024genagent}.} Metric: Resolve rate (\%). The table reports performance across three task types: Vanilla, Complex, and Creative, along with the overall average.}
    \label{tab:comfybench}
    \centering
    \begin{tabular}{l|ccc|c}
        \hline

        \hline
        
        \hline
        \textbf{Method} & \textbf{Vanilla} & \textbf{Complex} & \textbf{Creative} & \textbf{Total}\\ 
        \hline
        ComfyAgent~\cite{xue2024comfybench} & $46.00$ & $21.67$ & $15.00$ & $32.50$ \\
        HuggingGPT~\cite{shen2023hugginggptsolvingaitasks} & $21.00$ & $0.00$ & $5.00$ & $11.50$ \\
        LWM~\cite{liu2023world} & $24.00$ & $8.33$ & $5.00$ & $15.50$ \\
\rowcolor[gray]{.9}
\textbf{Ours} & \textbf{61.00} & \textbf{28.89} & \textbf{19.17} & \textbf{43.00} \\
\rowcolor[gray]{.9}
\textit{Error (Ours, 3 times)} & {\scriptsize $\pm5.00$} & {\scriptsize $\pm5.55$} & {\scriptsize $\pm3.33$} & {\scriptsize $\pm4.83$} \\

        \hline

        \hline

        \hline
    \end{tabular}
    \vspace{-0.2cm}
\end{table}

\paragraph{Performance on ComfyBench} Table~\ref{tab:comfybench} shows that our method achieves an overall success rate of \textbf{43.00\%} (±4.83), substantially outstripping the next-best approach, ComfyAgent, by over 10\% (\textbf{32.50\%}). In particular, our solution handles inherently non-atomic tasks like “merge model” (11 components, 11 links) and “image merge” (13 components, 17 links), which challenge other methods.

\begin{table}[t]
\centering


\setlength{\tabcolsep}{3pt}
\renewcommand\arraystretch{1.2}
    \caption{\textbf{Ablation study on inference design.} Metric: Resolve rate (\%). The table shows the effect of different inference components, evaluated under ComfyBench~\cite{xue2024genagent}.}
    \label{tab:ablation}
    \centering
    \begin{tabular}{cc|ccc|c}
        \hline

        \hline

        \hline
          \textbf{2-stage} & \textbf{refinement}& \textbf{Vanilla} & \textbf{Complex} & \textbf{Creative} & \textbf{Total}\\ 
        \hline
          & \ding{51} & $47.00$ & $10.00$ & $10.00$ & $28.50$  \\
         \ding{51} & & $32.00$ & $16.67$ & $5.00$ & $22.00$ \\
        \rowcolor[gray]{.9} 
         \ding{51} & \ding{51} & $\textbf{56.00}$ & $\textbf{28.33}$ & $\textbf{22.50}$ & $\textbf{41.00}$ \\
        \hline

        \hline 

        \hline
    \end{tabular}
    \vspace{-0.2cm}
\end{table}
\begin{figure}[t]
    \centering
\includegraphics[width=\linewidth]{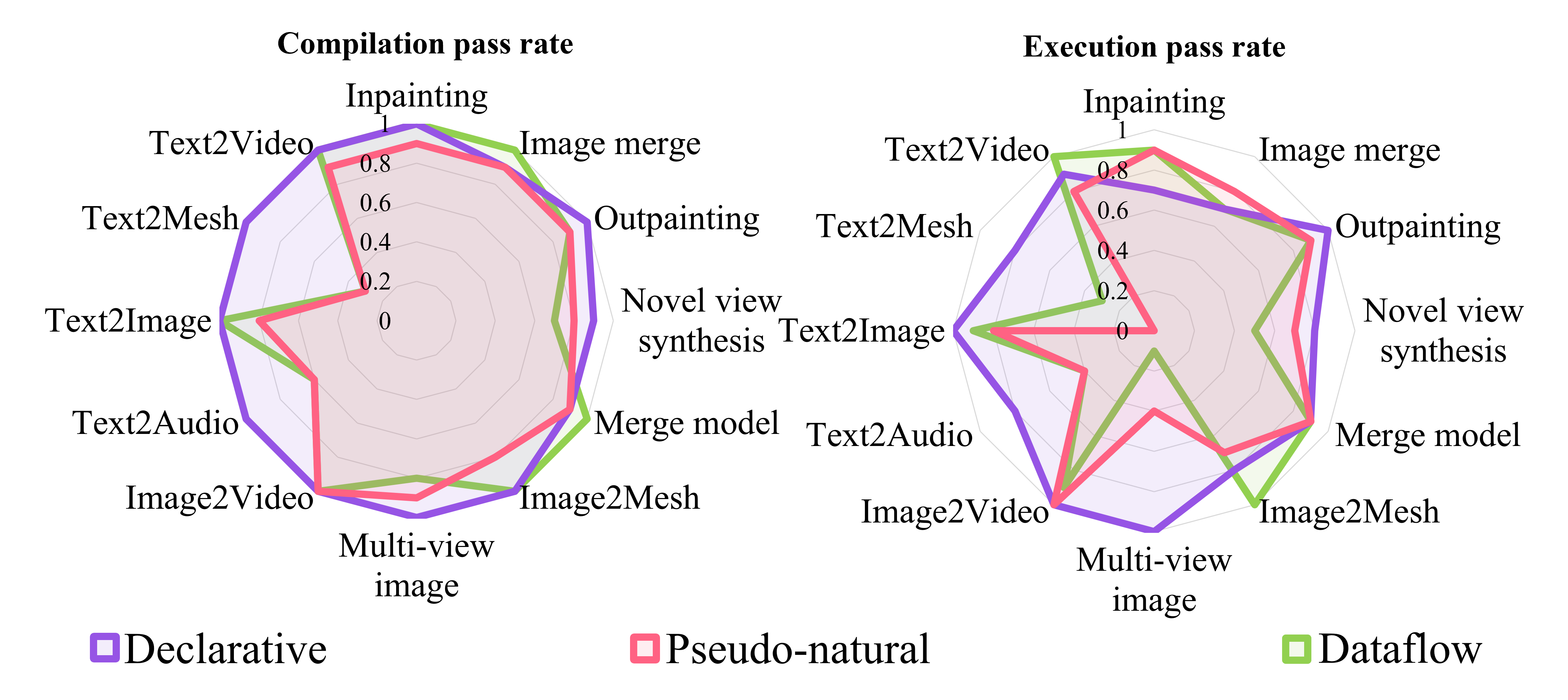}
    \caption{
\textbf{Comparison of syntax styles.} Metric:  Pass@$1$ (↑). See Appendix for details.
}
\vspace{-0.4cm}
    \label{fig:passrate}
\end{figure}

\paragraph{How to infer symbol flow?}
Table~\ref{tab:ablation} demonstrates the critical importance of our two-stage inference architecture (See Figure~\ref{fig:inference}) and iterative refinement mechanism (See Sec.~\ref{sec:inferring}). Removing either component significantly degrades performance across all task categories. With both components, our approach achieves a $41\%$ overall resolve rate on ComfyBench, compared to just $28.5\%$ with refinement alone and $22\%$ with two-stage generation alone.

\begin{table}[htbp]
\centering
\setlength{\tabcolsep}{14pt}
\renewcommand\arraystretch{1.2}
\caption{\textbf{Agentic design~\cite{xue2024genagent} \textit{vs.} symbolic inference (Ours) on our task suite.} We calculate the average pass rate (Pass@1, ↑) on compilation and execution on our $120$ task suite.}
\label{tab:agentic}
\centering
\begin{tabular}{l|cc}
\hline

\hline

\hline

Method & Compilation & Execution \\ \hline
GenAgent~\cite{xue2024genagent} & 0.84 & 0.63 \\
\rowcolor[gray]{.9} 
Ours & \textbf{0.98} & \textbf{0.87} \\ 
\hline

\hline

\end{tabular}
\label{tab:agentic_or_symbolic}
\vspace{-0.4cm}
\end{table}

\paragraph{Symbolic vs. agentic approaches}
As shown in Table~\ref{tab:agentic_or_symbolic} and Figure~\ref{fig:passrate}, our symbolic approach achieves higher success rates without the complexities of agentic designs. Unlike GenAgent~\cite{xue2024genagent}, which employs multi-step planning that can amplify errors and increase costs, our symbolic method maintains simplicity while minimizing error propagation. For straightforward tasks, this simpler approach leads to higher pass rates, though for more intricate workflows, integrating symbolic representations with agentic strategies may offer enhanced flexibility.

\begin{figure}[th]
    \centering
\includegraphics[width=\linewidth]{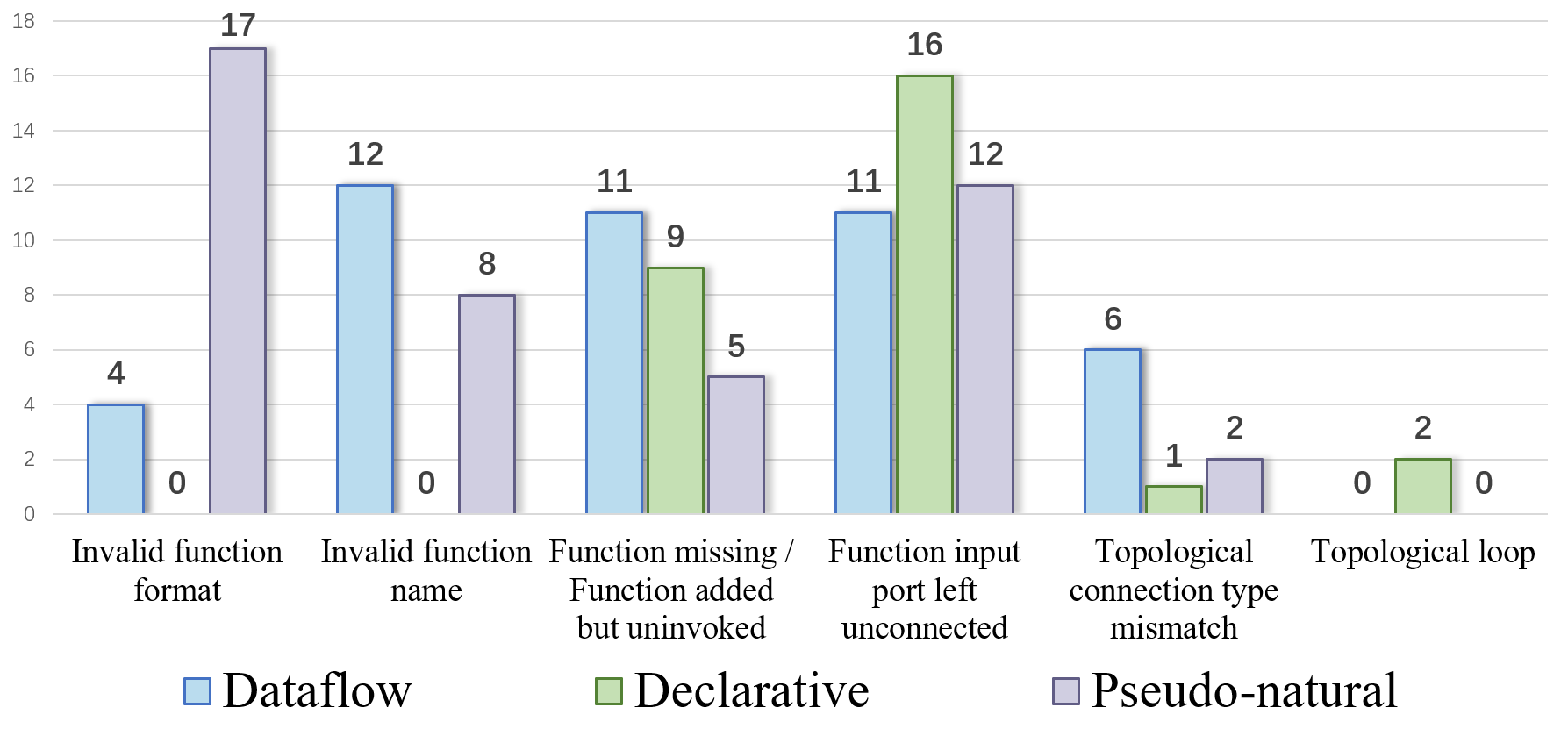}
    \caption{
\textbf{Comparative error distribution} for dataflow, declarative, and pseudo-natural syntax styles, illustrating six types of errors occur when testing on the $120$ generative tasks.
}
    \label{fig:error_counts}
    \vspace{-0.6cm}
\end{figure}

\begin{figure}[th]
    \centering
\includegraphics[width=\linewidth]{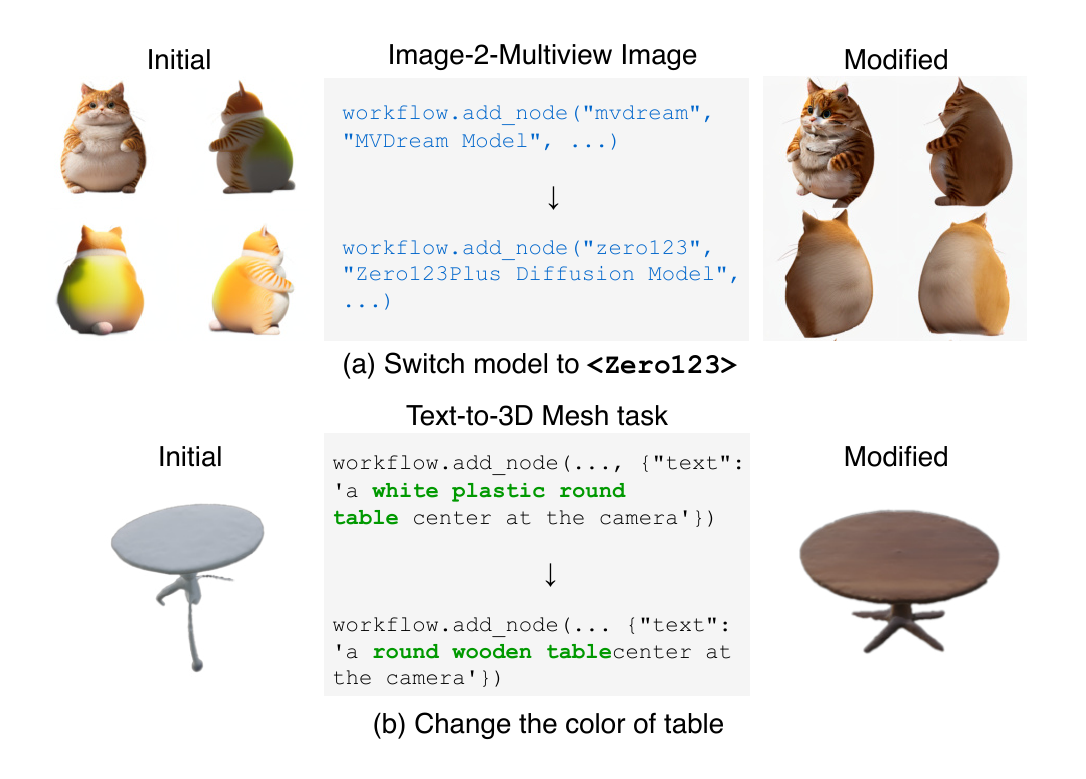}
    \caption{
\textbf{Symbolic Flow Editing.} We present examples of modifying (a) \textcolor{darkblue}{\textit{functions}}, where users can directly change models by editing code to achieve desired effects, and (b) \textcolor{darkgreen}{\textit{parameters}}, such as adjusting textual prompts (treated as a type of parameter) to alter the color of 3D assets.}
    \label{fig:edit}
    \vspace{-0.5cm}
\end{figure}

\paragraph{Representation: neural or symbolic?}
Our symbolic model outperforms neural models in task generality and output quality without additional training. Table~\ref{tab:quality_rank_comparison} highlights that our symbolic approach successfully handles all $120$ generative tasks, including complex categories such as 3D and video generation. In contrast, neural models are limited by their reliance on extensive training data, restricting their ability to manage diverse and complex tasks.

\paragraph{Explicit symbolic flow editing and error analysis}
Our symbolic representation enables precise control over distinct stages of generative tasks through explicit program modifications. Figure~\ref{fig:edit} illustrates examples of modifying \textcolor{darkblue}{\textit{\textbf{function}}} (model) and \textcolor{darkgreen}{\textit{\textbf{parameter}}} (textual prompt). Analysis of the $120$ test cases in Figure~\ref{fig:error_counts} reveals two key findings: \blackcircle{1} Higher readability in language design correlates with increased format errors, with pseudo-natural language formats showing more invalid code formats than dataflow or declarative styles. \blackcircle{2} Structurally rigid languages tend to introduce topological gaps and connection errors, suggesting that increased structural complexity challenges language models in maintaining accurate dependencies.
\section{Conclusion}
We have proposed a symbolic generative task description language, combined with an inference engine, providing a novel and efficient way to represent and execute multimodal tasks without the need for task-specific training. 
By leveraging a pre-trained large language model to infer symbolic task descriptions, our approach has successfully synthesized diverse multimodal tasks, demonstrating its flexibility and potential to unify different generative AI capabilities.
Our experiments on $120$ tasks and ComfyBench have shown that our framework has achieved performance comparable to unified multimodal models, highlighting its expandability and cost-effectiveness.

{
    \small
    \bibliographystyle{ieeenat_fullname}
    \bibliography{main}
}



\end{document}